# Promotheus: An End-to-End Machine Learning Framework for Optimizing Markdown in Online Fashion E-commerce

Eleanor Loh
eleanor.loh@asos.com
ASOS.com, London, United Kingdom

Jalaj Khandelwal
jalaj.khandelwal@asos.com
ASOS.com, London, United Kingdom

Brian Regan[1]
brian.regan@asos.com
ASOS.com, London, United Kingdom

Duncan A. Little
duncan.little@asos.com
ASOS.com, London, United Kingdom

## ABSTRACT

Managing discount promotional events ("markdown") is a significant part of running an e-commerce business, and inefficiencies here can significantly hamper a retailer's profitability. Traditional approaches for tackling this problem rely heavily on price elasticity modelling. However, the partial information nature of price elasticity modelling, together with the non-negotiable responsibility for protecting profitability, mean that machine learning practitioners must often go through great lengths to define strategies for measuring offline model quality. In the face of this, many retailers fall back on rule-based methods, thus forgoing significant gains in profitability that can be captured by machine learning. In this paper, we introduce two novel end-to-end markdown management systems for optimising markdown at different stages of a retailer's journey. The first system, "Ithax," enacts a rational supply-side pricing strategy without demand estimation, and can be usefully deployed as a "cold start" solution to collect markdown data while maintaining revenue control. The second system, "Promotheus," presents a full framework for markdown optimization with price elasticity. We describe in detail the specific modelling and validation procedures that, within our experience, have been crucial to building a system that performs robustly in the real world. Both markdown systems achieve superior profitability compared to decisions made by our experienced operations teams in a controlled online test, with improvements of 86% (Promotheus) and 79% (Ithax) relative to manual strategies. These systems have been deployed to manage markdown at ASOS.com, and both systems can be fruitfully deployed for price optimization across a wide variety of retail e-commerce settings.

1. Now at Spotify (brianr@spotify.com).

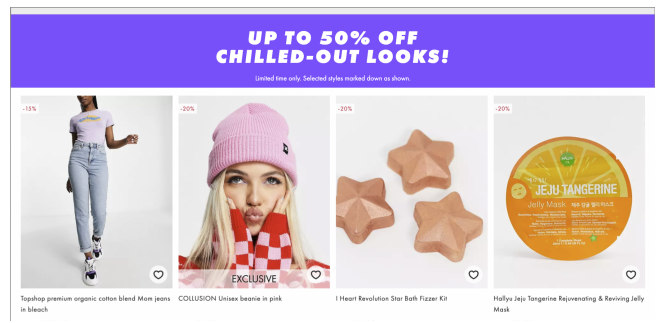

Figure 1: Promotional pricing at ASOS.com. ASOS is a major fashion e-commerce company, with >26M active customers worldwide and up to 90K products on its platform at any one time. Like other fashion e-commerce players, ASOS contends with fast-changing microtrends that can dramatically affect week-to-week demand for a given product. Markdown is a powerful tool for maintaining profitable operations at scale.

## 1 INTRODUCTION

Promotional pricing ("markdown") is crucial in retail, and especially in fashion, where retailers purchase stock far ahead of time, with little information about the trends and circumstances that will strongly modulate later demand [1, 2]. Markdown allows a retailer to clear stock in the face of imperfect stock buying decisions. In the fast-paced world of *online fashion e-commerce* in particular, markdown can often form a substantial portion of a retailer's operating budget [11, 19]. Efficient markdown management is thus crucial to building a sustainable, profitable business at scale (Figure 1).

Along with its importance to profitability, price optimization is also very challenging to get right [1–4, 11]. In the face of this difficulty, retailers often fall back on rule-based strategies decided by operational teams [11, 19]. Given the substantive domain expertise embodied in such operational teams, these manual decisions can be very effective; this has been the case at ASOS, which has maintained profitable operations over a long period of time. Nevertheless, we show here that there are powerful gains to be had from building a price optimization system with machine learning. This paper presents an end-to-end framework for managing markdown, including components that can be used to achieve different goals at different stages of the machine learning product's life-cycle (Figure 2). We make a number of important contributions: First, we present

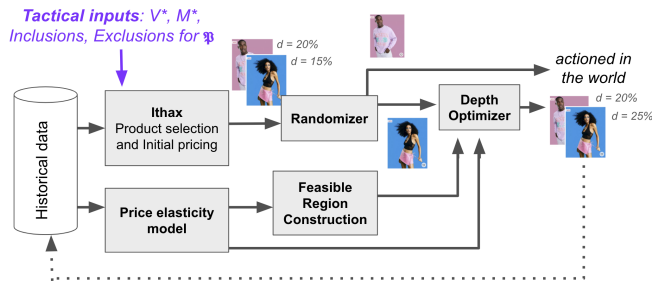

Figure 2: Overview of full markdown framework, Promotheus. Promotheus comprises multiple components. Firstly, "Ithax" is a supply-side markdown algorithm that selects products and sets initial depths. In order to measure incrementality and actively sample the action space more broadly, we randomly hold-out a proportion of selected products whose prices are set by Ithax (Randomizer). For all other products, depths are adjusted later on (Depth Optimizer) before the decisions are actioned in the world. We train a price elasticity model, and use offline validation to define the feasible region for decision making (Feasible Region Construction), as inputs to the Depth Optimizer. Although Ithax is a sub-process of Promotheus, it is itself a robust end-to-end markdown management system that can be deployed with financial control. Both Promotheus overall and Ithax specifically outperform manual pricing decisions in an online test, with Promotheus showing the best performance overall.

"Ithax", a supply-side algorithm for markdown optimization which selects and prices products for a promotional event, while keeping the overall markdown event within specific financial parameters. This algorithm enacts a rational pricing strategy from an inventory management perspective, and also serves as a powerful "cold start" algorithm that can be safely deployed to collect data for later demand modelling. While Ithax comprises just one part of our overall markdown management framework (Figure 2), it itself is a robust end-to-end solution, and we show that it delivers much stronger profitability compared to the manual pricing decisions made by our experienced operational teams. Next, we present "Promotheus", a price elasticity-based framework for markdown price optimization, that adjusts prices for products selected by Ithax, based on likely demand. We demonstrate the superior profitability of this framework compared to both the pricing decisions from Ithax and our operational teams. Importantly, we show these gains in an online real-world test, going beyond the offline numerical evaluation that is seen in many markdown proposals. Additionally, we describe the specific validation and modelling decisions that have allowed us to build a robust pricing system, and show how our framework has allowed us to deploy systems with robust decision quality. Finally, we address a key stumbling block for automated price optimization systems: the need for retailers to retain flexibility of operations, in order to shape the company's offering along qualitative, ad hoc dimensions that are imperfectly captured by any one machine learning framework [11]. The need for such tactical levers is crucial to a system's success in practice, but is rarely addressed in the literature. In fashion e-commerce, this manifests as a need to shape specific "themed" promotional events, either pinned to customer facing events (e.g. valentine's day), or to operational needs that have arisen on the supply side (e.g. the need to clear cold weather apparel due to overstocking decisions). We demonstrate how our core framework has been successfully adapted to provide these tactical levers, thus providing our operations teams with the ability to manage promotions with a powerful degree of tactical agility.

## 2 BACKGROUND

Price optimization is often framed as one of price elasticity: practitioners use models to predict sales as a function of price, often in combination with optimization methods (e.g. [6, 9, 10, 13]; see [12, 17] for survey). While demand prediction in price optimization presents superficially as a standard supervised learning problem, the crucial difference is in the usage of the models: in price optimization, we vary *price* as a decision variable, and are thus using models to generalise *beyond* the pricing policies expressed in historical data. In other words, demand prediction takes place in the context of partial information (rather than full supervision), and price optimization involves counterfactual inference: outcomes are unobserveable for prices other than those actioned historically, and demand curve learning inherently generalises across products with very different characteristics. It is therefore extremely important to evaluate price optimization systems using rigorous real-world tests, as opposed to offline numerical analysis [5].

*Validation under partial information* Validation in the context of partial information is challenging, since counterfactual outcomes are fundamentally unobservable in the historical data and thus unrepresented in offline metrics [14, 15]. Valid offline evaluation is thus a key challenge in designing price optimization systems. While many interesting and inventive markdown proposals exist [8, 12, 16, 18], these proposals often lack online, real-world evaluation, and rely on assumption-laden offline evaluation methods that leave one uncertain about incremental benefits in practice. We directly address this: in order to ensure that model degradations are observable, we (a) adopt specific procedures to measure offline model accuracy, (b) rely on online tests/model metrics as the gold standard for evaluation, and (c) adopt long-term randomised hold-outs, in order to (i) gather data on model accuracy across the action space, as well as (ii) measure the incremental impact of our price elasticity-based approach (relative to Ithax). Additionally, we outline a framework for empirically defining the decision space for price optimization, offline. This framework has allowed us to build a price optimization system whose online performance is stable, and whose profitability shows great advantage.

*Model specification and completeness* When supervised learning models are used within price optimization systems, pricing decisions are easily led astray by spurious correlations and outlier events in the historical data. One source of spurious correlation is the univariate relationship between (discounted) price and sales in the historical data: because deeper discounts are commonly allocated to poorer performing stock, supervised models learn that lower prices are associated with lower sales. This conflicts with the very strong expectation that price elasticity curves should be negative: expected sales are expected to increase (or be strictly non-decreasing), as prices decrease [4]. Demand curves that do not conform to such expectations are often inaccurate (outside of Veblen goods), and lack the interpretability that can be crucial to a pricing system's ability to gain stakeholder trust. As such, practitioners have often gone to extreme lengths to estimate models with negative elasticity (e.g.[6, 9]); we likewise adopt this concern and demonstrate that our demand models have the expected negative price-sales relationship, and discuss modelling decisions that were crucial to this outcome. More broadly, price elasticity models can be very sensitive to the details of model construction, in keeping with the causal inference axiom that treatment effects (i.e. causal impact of price changes) can only be reliably estimated within models whose functional form *accurately* and *completely* captures the structure of the world [7]. Within fashion, the assumption that models are *complete* (i.e. capturing all relevant demand-modulating covariates within the model structure) is particularly challenging, as sales can change dramatically in response to events that are difficult to capture completely in practice (e.g. social media virality). Although trends can in theory be captured with extensive data collection and feature engineering, it is often impractical to do so (e.g. social media could in theory be scraped to detect virality, but this is often prohibitively expensive). These extreme-valued outliers can dramatically skew predictions, but are unimportant within the context of markdown (whose job is to efficiently move the majority of stock, not capitalise on long-tail revenue opportunities). As such, we specify our model to de-emphasize the importance of such long tail events, in order to prioritise model stability overall.

*An end-to-end algorithm for "cold start" markdown optimization* A key contribution of this paper is in presenting "Ithax" as a "cold start" markdown optimization engine. This addresses a position that many retailers find themselves in: lacking a deep recorded history of price change data that can be used to build price optimization models. Although Ithax makes little attempt to consider demand, it still enacts a financially rational pricing strategy (in keeping with the inventory management literature [11]), and requires only information about a retailer's *current* stock/sales position in order to proceed. It thus fills an important niche: it constructs markdown events that conform to key financial parameters, without requiring historical data, thus allowing a retailer to preserve operational profitability while collecting data for later demand modelling. We demonstrate the superior performance of this algorithm relative to manual pricing decisions, and note that its simplicity and adaptability give it a wide applicability across many retail domains, including grocery, department stores and other e-commerce contexts.

## 3 SUPPLY-SIDE ALGORITHM ("ITHAX")

We describe "Ithax", a supply-side algorithm for managing markdown that (a) selects and prices products for a markdown event, while (b) keeping within financial constraints. It does so by allocating products to different discounts, evaluating a proxy for profitability of the overall stock pot, and repeatedly reallocating discounts based on whether this proxy is above or below target. Ithax was developed in close partnership with ASOS operations teams, and effectively embeds common rule-based logic [19] within a multi-objective algorithm inspired by binary search. We define the following constructs (see Table 1 for a worked example):

- **Discount depth** Let the discount depth $d_{p,t}$ be the proportion "percentage off" discount for product $p$ in time period $t$. We discretize depths $d_p \in D = \{0, 0.05, \ldots\} \in [0, 1]$, with the set of acceptable depths $D$ set by operations teams.

- **Inclusion set** Let $P_t$ denote the set of all products (selected from the product catalogue $\mathfrak{P}$) to be included in a specific markdown event at time t, with their specific allocated discount depths. A given product $p \in P_t$ iff its depth $d_{p,t} > 0$.

$$P_t = \{\forall p \mid d_{p,t} > 0\} \subset \mathfrak{P} \quad (1)$$

- **Stock Value** Let $V(P)$ denote stock value: the total "full price" value of a set of products $P$, computed as:

$$V(P_t) = \sum_p^{P_t} f_p \cdot k_{p,t} \quad (2)$$

where $f_p$ is the full price for the product $p$, and $k_{p,t}$ is the number of stock units at the start of time $t$, for product $p$.

- **Stock Depth** $M(P)$ denotes stock depth of a set of products $P$: the ratio of summed discounted-prices to summed full prices:

$$M(P_t) = 1 - \frac{\sum_p^{P_t}(1 - d_{p,t}) \cdot f_p \cdot k_{p,t}}{\sum_p^{P_t} f_p \cdot k_{p,t}} \quad (3)$$

- **Cover** Let $c_{p,t}$ denote cover for product $p$ in time period $t$: the number of weeks it would take for the product to completely sell out, if the number of units sold in each week future week were exactly as observed within time period t:

$$c_{p,t} = \frac{k_{p,t}}{s_{p,t}} \quad (4)$$

where $s_{p,t}$ is no. sold units in time period $t$ for product $p$

Ithax uses cover $c_p$ for $p \forall \mathfrak{P}$ to construct $P$ for a markdown event. Cover is a product-level metric that captures the rate of sale, and is widely used to measure retail product performance. Stock value and stock depth are high-level metrics that capture the intended sales and profitability, respectively: (a) higher stock values indicate more products being included in a promotional event, which is assumed to produce higher sales volume (b) higher stock depths indicate deeper discounting over the stock pot, indicating lower profitability. They are key algorithm inputs from the business that allow for tactical control over a single markdown event.

**Algorithm Aims** Ithax aims to construct $P_t^*$ for a markdown event at time $t$: a set of products $P$ (each with assigned discounts) to be included in markdown, whose stock value and stock depths conform to the targets. Ithax optimizes the following cost function:

$$f1(\hat{P}_t) = \frac{\mid V(\hat{P}_t) - V_t^* \mid}{V_t^*} \quad (5)$$

$$f2(\hat{P}_t) = \mid M(\hat{P}_t) - M_t^* \mid \quad (6)$$

$$argmin_{\hat{P}_t} \quad f1(\hat{P}_t), f2(\hat{P}_t) \quad \text{s.t. } r(c_{p,t}, d_{p,t}) > 0 \quad (7)$$

where $V_t^*$ and $M_t^*$ are the stock value and stock depth targets, and the constraint (a) applies only across the set of products $p \in \hat{P}_t$ (b) reflects the priority that poorer performing products (i.e. higher cover) should be assigned higher depths (latter part of eq. 7), iif they are included in the markdown event. This approach (assigning higher depths to poorer performing products) is common in retail, and we propose this as economically rational because, within a supply-side approach, markdown aims to sell out on *all* products. Higher cover thus indicates products that will need heavier discounts, in order to clear the current stock position (given current performance). We further expect that some products have such low demand that further discounts would not bring incremental sales ("zero-sellers"), and are thus not worth including in markdown given the limited space on the sale-related product listing pages. Thus, beyond some value of cover, products *should* not be included in markdown. As such, the algorithm takes as a starting point a linear mapping between cover and discount depth, where deeper depths map onto higher cover products, up to the point where zero-sellers are encountered (in Table 1 below, the threshold for "zero sellers" is 100 weeks of cover). It is the task of the algorithm to iteratively adjust this mapping (between cover and discount depth), such that stock depth and stock value targets are hit. In practice, we considered algorithm solutions to be of sufficient quality for time period $t$ ( i.e., $P_t = P_t^*$) when the following apply:

$$\hat{P}_t = P_t^*, \iff [f1(\hat{P}_t) < 0.05, f2(\hat{P}_t) < 0.005] \qquad (8)$$

Many solutions $P_t^*$ may exist for a given $(V_t^*, M_t^*)$, and we consider all solutions to be equally good. See 6.1.1 for tactical levers that provide operational control over concerns beyond $V_t^*, M_t^*$.

### 3.1 Algorithm

At a high level, Ithax (a) starts with a set of rules for allocating products to depths according to their cover (b) evaluates the proposed products and depths with respect to stock depth (a proxy for profitability), and (c) iteratively adjust the cover-based rules to reallocate products, based on whether the current stock depth is above or below target. Formally, Ithax aims to select individual products to include in a markdown event, and assigns depths $d_p$ for $p \forall P \subset \mathfrak{P}$. It does so by taking as a starting point a mapping between cover bands and a fixed set of discount depths, $B_0$, and then iteratively adjusting this mapping (effectively reselecting products and reallocating $d_p$ for $p \forall P$) to hit $V*$ and $M*$. It uses binary search to identify an optimal way of partitioning all products by cover, into different cover "bands" with specific allocated depths.

Let $B_i$ denote a specific mapping between cover bands and discount depths, at iteration $i$ of Ithax's workflow. Table 1 shows an example $B_i$: the range of potential cover values for a candidate $\hat{P}$ is partitioned into different "cover bands", denoted $b$. Each band $b \in B$ consists of minimum/maximum cover "bounds", ($c_b^{min}, c_b^{max}$), and maps onto a single unique discount depth value $d_b$, where any product whose cover falls within the band's range is then assigned the corresponding depth $d_b$. Within $B$, cover bands are ordered sequentially, from lowest cover to highest cover, with no gaps in cover in between adjacent cover bands (i.e. a band's upper bound is also the lower bound of the adjacent, higher cover band). Higher cover is associated with higher discount depths; however, beyond some value of cover, products are no longer selected for markdown (e.g. products where $c_p$ > 100, in Table 1). Identifying this maximum cover value for the products in $P$ is one of the tasks of the algorithm.

Ithax involves two separate, alternating sub-processes (Algorithm 1): a depth allocation step, and a boundary adjustment step. The depth allocation step constructs a $\hat{P}$: using the candidate cover-depth mapping $B_i$ (i.e. currently under consideration in iteration $i$), and selects products to hit $V^*$ (eq. 5). The boundary adjustment step then evaluates $M(\hat{P})$ with respect to $M^*$ (eq. 6), and proposes a new cover-depth mapping $B_{i+1}$ for the next iteration $i + 1$, based on whether $M(\hat{P})$ is higher or lower than $M^*$. Note:

| Cover Band | Boundaries $(c_{band}^{min}, c_{band}^{max}]$ | Discount Depth |
|---|---|---|
| i | (0, 20] | 0% |
| ii | (20, 40] | 15% |
| iii | (40, 60] | 30% |
| iv | (60, 70] | 50% |
| v | (70, 100] | 75% |
| vi | (100, ∞] | 0% |

**Table 1: Example mapping between cover bands and discount depths, $B$. Here, markeddown products $P$ consists of all products where $20 < c_p <= 100$. The bands with the highest and lowest discount depths ($b_i^{highest}, b_i^{lowest}$, respectively) are bands (v) and (ii) respectively. Note that products in cover bands (i) and (vi) $\notin P$, because $d_p$=0 for these products (eq. 1). We refer to a band's $width_b = c_b^{max} - c_b^{min}$.**

- Within each iteration of Ithax, only one cover band ("$b_x$") is selected for adjustment within the boundary adjustment step.
- In Algorithms 1-4, $i$ refers to an iteration within Algorithm 1. See Table 1 for an example starting set of cover-depth mapping $B_0$, and Appendix Section 8.6 for notes re constructing $B_0$.

---

**Algorithm 1** Overall logic of product selection algorithm, Ithax

---

**Input:** $V^*, M^*, B_0$
**Output:** $\hat{P} = P^*$
1: Initialise $b_x := b_i^{highest}$
2: **for** i = 1,2 … **do**
3:     Construct $\hat{P}_i$ via Algorithm 2 and compute $V(\hat{P}_i), M(\hat{P}_i)$
4:     **if** $\hat{P}_i = P^*$ **then**         ▷ Evaluated using eq 8
5:         **return** $\hat{P}$
6:     **else**
7:         **if** $M(\hat{P}_i) > M^*$ **then**
8:             $b_x := b_i^{highest}$
9:             $(B_{i+1}, b_x) := \text{adjust}(B_i, b_x)$    ▷ Using Algorithm 3
10:        **else**
11:            $(B_{i+1}, b_x) := \text{adjust}(B_i, b_x, M(\hat{P}_i), M(\hat{P_{i-1}}), i)$    ▷ Using Algorithm 4
12:        **end if**
13:    **end if**
14: **end for**

---

*3.1.1 Depth allocation step.* The depth allocation step (Algorithm 2) implements the (cover band, depth) mapping $B_i$ of iteration $i$, selecting products $P$ s.t. $V(\hat{P}) = V^*$ (ignoring stock depth). The only "decision making" implemented here is: where the inclusion of all products in a cover band $b$ would lead to $V(\hat{P}) > V^*$), this sub-process adds a subset of products to allow $V(\hat{P})$ to approach $V^*$ as closely as possible without exceeding it (line 5).

The depth allocation step adds products from a single band at a time, starting with the highest band ($b_i^{highest}$; e.g. in Table 1, $b_i^{highest}$=band (v)) and moving to lower bands iteratively. This expresses the priority given to worse performing (i.e. higher cover) products, in deciding which products to include in markdown.

**Algorithm 2** Depth Allocation Step populates $\hat{P}$ according to $B_i$

**Input:** $V^*$, $B_i$
**Output:** $\hat{P}$
1: Initialise $\hat{P} = \{\}$ (to be filled with (product, depth) tuples)
2: Identify $b_i^{highest}$, $b_i^{lowest}$: highest/lowest cover band by discount depth (depth > 0)
3: **for** b in $[b_i^{highest}, b_i^{highest-1} \ldots b_i^{lowest}]$ **do**
4:     Calculate stock value of products in band $b$, $V(P_b)$.
5:     **if** $V(\hat{P} \cup P_b) \le V^*$ **then**
6:         $\hat{P} := \hat{P} \cup P_b$
7:     **else**
8:         Randomly select subset $P_b^x \subset P_b$ to roughly achieve:
9:         $argmax_{P_b^x} \mid P_b^x \mid s.t.(V(P_b^x) \le V^* - V(\hat{P}))$
10:        $\hat{P} := \hat{P} \cup P_b^x$
11:    **end if**
12: **end for**

It is possible in theory that depth allocation might terminate without $V(\hat{P}) \approx V^*$ (i.e. without satisfying $f1 < 0.05$; eq. 5), if there are not enough products in the cover ranges to fulfil the target stock value $V^*$. We implicitly assume that $V^*$ is reasonably scaled with respect to the available catalogue (see Appendix Section 8.6).

*3.1.2 Boundary Adjustment Step.* After each candidate $P_i$ is constructed (by Algorithm 2), Ithax evaluates $P_i$ against $M^*$ (in Algorithm 1). If convergence criteria has not been reached, the boundary adjustment step then alters the boundaries in $B_i$, thus changing the products that are allocated to each depth. Depending on whether $M(\hat{P}_i)$ is >/< than $M^*$, different subroutines are triggered. In both subroutines, we (a) identify target band $b_x$ within $B_i$, whose upper bound is to be adjusted, (b) increase/decrease the size of the target $b_x$ via binary cut (c) adjust all other boundaries within $B_i$ so as to maintain the 'width' of all other cover bands other than $b_x$.

*3.1.3 Target Stock Depth is exceeded [ $M(\hat{P}_i) > M^*$ ].* When $M(\hat{P}_i) > M^*$ occurs, Ithax reduces the number of products that are assigned higher depths, by removing products from $\hat{P}_i$ or by allocating some products to lower depths (Algorithm 3). It applies a binary search reduction to identified target band $b_x$, which is initialised to be band with the highest depth (denoted $b_i^{highest}$; $b_i^{highest}$ = band (v), in Table 1). Note that all cover band boundaries $c_{band}^{max/min}$ refer to elements of $B_i$ (see Table 1 legend for detail), and $B_i$ is thus updated to $B_{i+1}$ by the time it is returned to Algorithm 1. We note:
- Before the target band $b_x$ is adjusted, Ithax first evaluates if it is adjustable. A band is adjustable iif its depth > 0, and $width_{band}$ >threshold (threshold=3; see Appendix Algorithm 5). If a band is not adjustable, consideration moves to the next band, moving from higher-cover to lower-cover bands (i.e. in Table 1, $b_x$ moves from (v) → (ii)); this repeats until an adjustable band is found (see Appendix Section 8.6 for edge case handling).
- Algorithm 3 effectively reduces $M(\hat{P}_i)$ by reducing width of $b_x$ (allocating fewer products to $d_{b_x}$), and moving all higher bands downwards to maintain their width. This removes products from $\hat{P}_i$ and decrements $c_{b^{highest}}^{max}$. For intuition, see Appendix Section 8.3 for a worked example of Algorithm 3.

**Algorithm 3** Adjusting $B_i$ for $M(P_i) > M^*$

**Input:** $B_i$, where $i$ indexes the iteration within Algorithm 1
**Input:** $b_x$: target band for boundary adjustment
**Output:** $B_{i+1}$: updated $B$ for iteration $i+1$ in Algorithm 1
**Output:** $b_x$: band to be adjusted in iteration $i+1$ in Algorithm 1
1: **while not** is_adjustable($b_x$) **do**    ▷ (Appendix Algorithm 5)
2:     $b_x := b_x - 1$
3: **end while**
4: Initialise x = $(\frac{c_{b_x}^{max} - c_{b_x}^{min}}{2})$
5: $c_{b_x}^{max} := c_{b_x}^{max} - x$
6: **for** b in $[b_x + 1, b_x + 2 \ldots b_i^{highest}]$ **do**
7:     $c_{b_x}^{min} := c_{b_x}^{min} - x$
8:     $c_{b_x}^{max} := c_{b_x}^{max} - x$
9: **end for**
10: **return** $B_i, b_x$       ▷ $B_i$ has been updated, as $B_{i+1}$

*3.1.4 Candidate Stock Depth is insufficiently high [ $M(\hat{P}_i) < M^*$ ].* Where $M(\hat{P}_i) < M^*$, Ithax must either (Option A) add new products to $P_i$, or (Option B) transfer products from a lower to higher depth band. Although there are some edge cases to consider, the core logic of this subroutine (line 16 onwards) is the mirror image of that shown in Algorithm 3, shown in Algorithm 4. For clarity, we again provide a concrete example in Appendix Section 8.4.

In Option A, we increase $M(\hat{P}_i)$ is by widening $b_i^{highest}$, adding products to $P_i$ at a high depth. This is the "first" action attempted by Algorithm 4 (lines 2, 8; note lines 1-3 define an edge case where i=1 and $M(\hat{P_{i-1}})$ is undefined). The only situation in which widening $b_i^{highest}$ is unlikely to be effective is when there are few products with $c_p > c_{b^{highest}}^{max}$. This is detected by noticing that $M(\hat{P_{i-1}})$ has not changed relative to the previous iteration (line 6), and thus moving to consider lower bands (i.e. Option B). In Option B: Algorithm 4 increases $M(\hat{P_{i-1}})$ by decreasing the width of $b_x$ (by magnitude $x$), making the highest-depth band wider (by that same magnitude $x$), and maintaining the width of all bands in between. This effectively transfers products from a lower depth to a higher one.

*3.1.5 Convergence and specific assumptions.* The boundary adjustment algorithm works via binary search: with each iteration of boundary adjustment, the "size" of the adjustment is 50% of the target cover band's size. Together, Algorithm 3 and Algorithm 4 allow Ithax to iteratively approach the target stock depth, terminating when the observed deviation is acceptably small (see Appendix Sections 8.5-8.6 for considerations re convergence).

## 4 PROMETHEUS: FULL PRICE OPTIMIZATION

The supply-side algorithm Ithax constructs reasonable markdowns without estimating demand. While it has powerful applicability (see Section 2), it relies on *indirect* proxies for profit and sales volume (stock depth and stock value, respectively). It is also more likely to spend markdown budget on products where incremental discounts are unlikely to realise incremental profit gains. For this reason, it is still preferable to employ demand estimation in our full price optimization framework, moving away from inventory management and towards profit optimization. In a world in which we have *direct* values for (expected) sales and profit (e.g. via demand

**Algorithm 4** Adjusting $B_i$ for $M(P_i) < M^*$

**Input:** $B_i, b_x, M(\hat{P_i}), M(\hat{P_{i-1}}), i$
**Output:** $B_{i+1}, b_x$

1: **if** i == 1 **then**
2: $\quad c_{b_x}^{max} := c_{b_x}^{max} + \frac{c_{b_x}^{max} - c_{b_x}^{min}}{2}$
3: $\quad$ **return** $B_i, b_x$ $\quad\quad\quad\quad\quad\triangleright B_i$ has been updated, as $B_{i+1}$
4: **else**
5: $\quad$ **if** $b_x == b_i^{highest}$ **then**
6: $\quad\quad$ **if** $M(\hat{P_i}) \approx M(\hat{P_{i-1}})$ **then** $b_x := b_x - 1$
7: $\quad\quad$ **else**
8: $\quad\quad\quad c_{b_x}^{max} := c_{b_x}^{max} + \frac{c_{b_x}^{max} - c_{b_x}^{min}}{2}$
9: $\quad\quad\quad$ **return** $B_i, b_x$ $\quad\triangleright B_i$ has been updated, as $B_{i+1}$
10: $\quad\quad$ **end if**
11: $\quad$ **end if**
12: $\quad$ **while not** is_adjustable($b_x$) **do**
13: $\quad\quad b_x := b_x - 1$
14: $\quad$ **end while**
15: **end if**
16: Initialise x = $(\frac{c_{b_x}^{max} - c_{b_x}^{min}}{2})$
17: $c_{b_x}^{max} := c_{b_x}^{max} - x$
18: **for** b in $[b_x + 1, b_x + 2 \ldots b_i^{highest}]$ **do**
19: $\quad c_{b_x}^{min} := c_{b_x}^{min} - x$
20: $\quad$ **if** b < $b_i^{highest}$ **then**
21: $\quad\quad c_{b_x}^{max} := c_{b_x}^{max} - x$
22: $\quad$ **end if**
23: **end for**
24: **return** $B_i, b_x$ $\quad\quad\quad\quad\triangleright B_i$ has been updated, as $B_{i+1}$

forecasting), we are able to optimise directly for those expected outcomes, and stock value/depth as lose their importance.

Figure 2 shows the full markdown optimization framework, with individual components described below. In this framework, we continue to rely on Ithax to select products to include in the markdown event ($\hat{P_t}$). We then use a model to forecast outcomes across the action space $d_p \in D_p$ for $p \forall P$. We use offline validation to define the action space for each individual product $D_p$. Finally, we optimise depths on a product level by selecting $d_p^* \in D_p$ to maximise expected outcomes according to business objectives.

### 4.1 Demand forecasting model

We model demand for products independently, ignoring cannibalization. Specifically, we predict log-sales $s_{p,t}$ for product $p$ at time $t$, as a function of its depth $d_{p,t}$ and a covariate vector $x_{p,t}$:

$$log(s_{p,t}) = f(d_{p,t}, x_{p,t}) \quad (9)$$

**Learning algorithms** We considered linear models, tree-based models, and neural nets in our search space. We ultimately employed gradient-boosted trees as our model, using the lightgbm python library and applying a monotonic constraint to the *depth* decision variable so as to avoid the model learning spurious negative relationships between depth and forecast volumes. The covariate $x_{p,t}$ included many features related to product characteristics (n=25 features), historical price status, recent product performance in different sales channels (e.g. full price vs markdown), and seasonality. We were able to confirm with later analysis that this learning algorithm produced negative price elasticity curves, as expected.

**Extending training history** Because seasonality strongly modulates product demand in the fashion sector, we also include several seasonality covariates, and train our models with several years of historical data to expose the model to several epochs of a seasonal trend. Although this allowed our models to learn long-range seasonal patterns, this initially did not improve our forecast performance - potentially because this also increased the model's exposure to rare, long-tailed that are inescapable in the fashion (see Section 2 for an example). These events may be particularly destabilising for the tree-based models, since such outlying data-points effectively skew the entire leaf's estimate upwards. In order to counteract this, we applied winsorization to the target variable $s_p$ in the training data, at the 0.5%-ile. Adding winsorization allowed us to realise the expected performance improvement from adding extensive historical data (along with seasonality feature engineering) to our models. Importantly, while we winsorized the target for prediction, we continued to validate our models based on their ability to predict un-winsorized (log-transformed) sales. We expected to see worse prediction performance for these long tail events, in exchange for improved performance for other data points.

### 4.2 Validation

Our model metric was $WAPE = \frac{\sum_i^N |\hat{s_i} - s_i|}{\sum_i^N s_i}$. Additionally, we designed validation with the partial information problem in mind: training error may not be reflective of the error to be observed once a model is deployed, because future chosen actions may overlap poorly with the actions in the historical data, for certain product groups. In order to measure the extent to which the partial information problem can be overcome (given the generative process that gave rise to the historical data), we adopted time-series K-Fold cross validation procedure: for a single fold, the latest 5 weeks are held out as the out-of-sample validation set, with earlier historical data used for model training. Additionally, we ran K=10 training/validation cross-folds, with each additional fold going backwards in time by 5 weeks. Thus we effectively evaluate models according to their ability to accurately predict events going forward in time, where (a) the partial information problem is always present in the training/validation split (b) the out-of-sample validation set covers one full retail season. Adopting this specific validation procedure allowed us to build models that were stable across extended periods of time: before we adopted this specific procedure, we frequently found that model performance varied wildly over time, despite performing favourably during single point-in-time modelling exercises.

## 5 DEPTH OPTIMIZATION

### 5.1 Feasible region construction

We select the optimal depth $d_{p,t}^* \in D_{p,t}$ at time $t$, for all products $p \forall P$. We defined the feasible region for $D_p$ by evaluating forecast performance using the specific time-series KFold described above, computing out-of-sample WAPE and aggregating across internal product groupings (Table 2). This allowed us to estimate forecast accuracy, for different depths in different product groups, knowing that offline product-level WAPE would always encounter the partial information problem. We then used forecast accuracy to limit the action space to regions where the model was acceptably good. Thresholds were determined based on early tests; these thresholds can also be determined by Monte Carlo sampling over the error distribution to estimate the business impact of model inaccuracy.

At this point, the feasible region may be too limited to be practically useful. In such cases, this exercise serves to guide model development, exposing where exploration or other measures are necessary. In practice, our extensively developed models produced large feasible regions. We propose this procedure as a practical, principled way of constraining decision spaces given the inherent partial information problem in price elasticity modelling, since the extent to which forecast accuracy will generalise to new decision spaces will always be difficult to determine a priori.

| Depth (norm. units) | Group A | Group B | Group C | Group D |
|---|---|---|---|---|
| 0.2 | 0.57 | 0.579 | 0.397 | 0.844 |
| 0.3 | 0.460 | 0.453 | 0.474 | 0.449 |
| 0.4 | 0.458 | 0.466 | 0.496 | 0.441 |
| 0.5 | 0.450 | 0.507 | 0.505 | 0.445 |
| 0.6 | 0.496 | 0.526 | 0.540 | 0.485 |
| 0.7 | 0.505 | 0.509 | 0.566 | 0.461 |
| 0.8 | 0.641 | 0.632 | 0.626 | 0.537 |

Table 2: Feasible region, constructed offline with time-series KFold. WAPE is shown by depth for 4 product groups, with infeasible regions highlighted in grey (threshold=0.55). For commercial sensitivity, depths are scaled to arbitrary units.

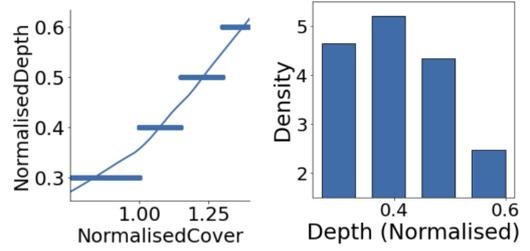

Figure 3: Constructed solutions $P^*$. *Left*: Relationship between cover/depth in Markdown Event A. Ithax allocates deeper depths to products with higher cover (indicating poorer performance). All values are scaled to arbitrary units for commercial sensitivity. *Right*: Density of products assigned to different discount depths for Markdown Event A. Allowable depths for each event are set by operations teams.

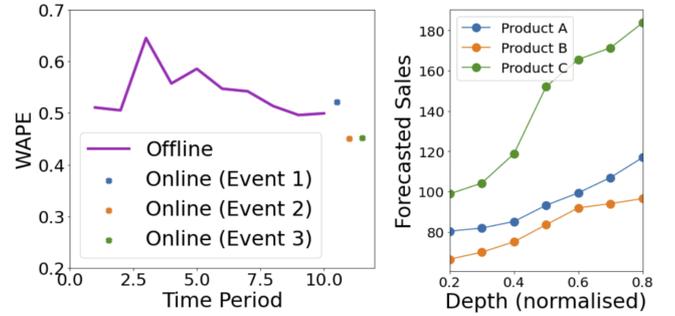

Figure 4: Performance of demand forecasting model. *Left*: Offline and Online model WAPE. Offline model performance data was fetched from our (deployed) model training pipeline, which automatically scores offline model performance using time-series KFold going back one full year. Dots show online model performance for three different markdown events, spanning over 1 month (as predicted by the same model whose offline performance is shown). *Right*: Price elasticity curves for 3 random products. We generated forecasts across a range of depths for all products in the catalogue, to evaluate if price elasticity curves were strictly non-decreasing. As shown, price elasticity can vary significantly from product to product, but is strictly non-decreasing in all cases.

## 5.2 Decision making

After defining the feasible region $D_p$ for each product group, we optimised discount depths $d_p$ on an individual product level, ignoring cannibalization and defining the optimal depth $d_p^*$ as:

$$d_p^* = argmax_{d_p \in D_p} \hat{s_{p,d}} \cdot \hat{g_{p,d}} \quad (10)$$

where $\hat{g_{p,d}}$ is the total expected profit $g$ for product $p$ at depth $d$, (computed by combining forecast $\hat{s_{p,d}}$ with product-level unit profitability). This objective function optimises for the product of expected profit and expected sales, reflecting the dual priorities of profitability and clearance inherent in markdown management.

## 6 RESULTS
## 6.1 Supply-side Markdown Algorithm

**Constructed solutions** $P^*$ We evaluated Ithax's performance using genuine targets, for multiple markdown events spanning >3 months. Ithax successfully hit stock depth and stock value targets for every single markdown event (Table 3), always converging to $P_t^*$ within 25 iterations (wall time <30 minutes; Appendix Figure 5). The range of stock value and stock depth targets for the events shown is representative of typical targets from across a full retail season. Products with poorer performance (i.e. higher cover) were assigned deeper discount depths by Ithax, as is consistent with a supply-driven goal to sell out all products over time (Figure 3, left). Within each markdown event, Ithax also allocated more products to middling depths, compared to extreme ones (Figure 3, right). This was a desirable quality, for our operations teams.

*6.1.1 Extensions for tactical agility.* We extended the algorithm to further enable operations teams to "shape" the contents of a promotional event, in service of other qualitative goals. Specifically:
**Inclusions and exclusions** We adapted Ithax to allow operations teams to force specific products into markdown ("Inclusions"), as well as exclude portions of the catalogue from consideration ("Exclusions"). These are trivially accomplished by (a) initializing $\hat{P}$ to be a non-empty set within Algorithm 2 (Inclusions), (b) excluding products from $\mathfrak{P}$ (Exclusions). We note that Ithax was able to handle the usage of such levers even when they were pushed to extremes (in a minority of cases): Exclusions have sometimes removed >90%

| Markdown Event | Stock Depth: Target (arbitrary units) | Stock Depth: Achieved (arbitrary units) | Stock Depth: % Discrepancy | Stock Value: Target (arbitrary units) | Stock Value: Achieved (arbitrary units) | Stock Value: % Discrepancy |
|---|---|---|---|---|---|---|
| A | 0.444 | 0.446 | -0.45% | 0.875 | 0.8749 | 0.02% |
| B | 0.5 | 0.508 | -1.60% | 1.5 | 1.4991 | 0.06% |
| C | 0.5 | 0.499 | 0.12% | 1.475 | 1.4746 | 0.03% |
| D | 0.46 | 0.468 | -1.74% | 1.375 | 1.3745 | 0.04% |

**Table 3: Product selection algorithm successfully hits financial targets. Financial targets and achieved values (in arbitrary units) shown for four different markdown events, spanning a 3 month period. Stock depth and stock value targets were successfully hit by the algorithm (discrepancy always <2% relative). Note that for reasons of commercial sensitivity, stock value and stock depth numbers (targets and achieved values) have been scaled to arbitrary**

of the catalogue, whereas Inclusions were occasionally specified such that the specific products had aggressively high stock depths, used alongside overall stock depth targets that were aggressively low (e.g. M($P_{inclusions}$) - M($P^*$) >10%). In all cases, Ithax successfully hitting overall stock depth and stock value targets, without any performance degradation.

**Product group priority** To enable markdown to have product group emphases, we adapted Ithax to allow operations teams to allocate different portions of the overall stock value target to different product types (i.e. splitting $V^*$ into $[V^*_{Group1}, V^*_{Group2}, ...V^*_{Groupr}]$). We enabled this by adapting Ithax to run Algorithm 2 separately for each product group (i.e. effectively calling Algorithm 2 r times in line 3 of Algorithm 1, instead of once). This successfully allowed the algorithm to hit these group-wise stock value targets (Table 4), with no degradation to performance on stock depth.

| Product Group | $V^*$ Target (norm. units) | $V$ Achieved (norm. units) | % Discrepency |
|---|---|---|---|
| 1 | 0.0960 | 0.0960 | 0.01% |
| 2 | 0.2015 | 0.2015 | 0.02% |
| 3 | 0.1330 | 0.1329 | 0.09% |
| 4 | 0.4445 | 0.4444 | 0.02% |

**Table 4: Product group prioritisation. Ithax was adapted to enable product prioritization via the specification of group-level $V^*$ targets. The table below shows the requested product group prioritisation (stock value targets per product group; arbitrary units) that were used in Markdown Event A in Table 3. Ithax successfully hit group-level $V^*$ targets, while maintaining the $M^*$ across the entire markdown event.**

### 6.2 Demand forecasting
**Offline/Online WAPE** Our best model was a lightgbm model (WAPE≈0.53; best regression≈0.68, neural nets≈0.71). Figure 4 shows offline and online WAPE covering a full year, where online WAPEs are shown for the *same* model as was used for markdown events in periods $t = 11 - 12$. Online WAPE was reasonably similar to offline.
**Price elasticity is strictly negative** We also verified that price elasticity was negative, by examining forecasted outcomes for each product across a range of depths. Although we applied monotone constraints to the lightgbm on depth, this does not guarantee that price elasticity will be negative due to spurious correlations. We avoid this unwanted outcome here: for 100% of products in the catalogue, forecasted sales were strictly non-decreasing as the price decreased across the entire forecasted range. Within the feasible region of considered discount depths, forecasted sales strictly increased 99% of the time on average. Figure 4 right shows the price elasticity curves for 3 random products in the catalogue.

### 6.3 Business KPIs
**Online test** We evaluated our markdown systems via an online test, run in all available markets and sales channels. Half the stock value budget was used by the operations team to manually price products for promotion ("Manual" condition), while the other half was used for algorithmic price optimization. Because Promotheus relies on Ithax for product selection, we used Ithax to select products for the markdown event (for the AI portion of the stock value budget). After product selection, we randomly assigned 50% of products to have their prices further adjusted by Promotheus (i.e. using the price elasticity and depth optimization procedure; "Full Optimization" condition). The remaining 50% of products retained the prices as decided by Ithax ("Supply-side" condition).

Stock depths were equivalent for manual vs algorithmic markdown decisions. We measured profit generated per product, which was analysed using non-parametric tests due to the data being not normally distributed (p<0.001). The test condition significantly influenced profitability (H=25.8 p<0.001), with the highest profitability in the Full Optimization condition, followed by the Supply-Side condition, followed by the Manual condition. Pairwise uplift comparisons confirm the relative superiority of the supply-side algorithm over manual pricing decisions, with further incremental gains from full price optimization (Table 5; note medians reported in keeping with the use of non-parametric tests). These results validate the usefulness of our markdown systems in helping to optimise markdown operations in a fashion e-commerce context.
**Deployment** The full algorithm has been deployed and is actively used to serve markdown events, with Ithax continuing to play a crucial role in product selection (Figure 2). Additionally, we use Ithax to serve a holdout proportion of products, which allows us to ongoingly monitor model and decision quality, by enabling live, always-on estimates of the incrementality of our price elasticity-based algorithm over a simpler supply-side approach.

|  | Uplift of Medians | Uplift of Means | P Value |
|---|---|---|---|
| Full Optimisation > Supply-side | 4.10% | 2.80% | $p < 0.05$ |
| Full Optimisation > Manual | 86.60% | 31.30% | $p < 0.001$ |
| Supply-side > Manual | 79.30% | 27.70% | $p < 0.001$ |

Table 5: Pairwise comparison of profitability from online test. P values are reported using the Mann Whitney U test.

## 7 CONCLUSIONS

In this paper, we present two solutions for managing markdown, which are together deployed to manage markdown at ASOS.com. Our solution includes a supply-side markdown algorithm that enacts a rational pricing policy without demand estimation, as well as a full framework which uses price elasticity to optimise for profit and clearance goals. Both approaches show vast superiority in profitability, even compared to experienced operational teams. Importantly, we demonstrate these gains in a rigorous real-world test, going beyond the offline numerical analysis that is common (but fundamentally limited) in the literature. In addition to demonstrating the effectiveness of this approach, we provide guidance about crafting validation and policy spaces to build robust price optimization, and outline how our system has been adapted to provide operational teams with a high degree of tactical agility. This adaptability to real world tactical concerns is crucial for real world systems, and is also often lacking in other price optimization proposals, however inventive. Our demand estimation model produces interpretable price elasticity curves (i.e. with the expected negative relationship between price and sales), where offline estimates of WAPE line up with online, on-policy performance after deployed. We note that our proposed solution is also easily implemented using open-source technologies (python, lightgbm library). We demonstrate both (a) the superiority of our two algorithmic solutions over the legacy, manual markdown management process, as well as (b) the further incremental gain of our price elasticity-based approach compared to the supply side algorithm. We further highlight the usefulness of the supply-side algorithm in providing a "cold start" solution to price optimization, which can be useful for bootstrapping data collection for later price elasticity modelling. Lastly, we note that the use of simple retail metrics, and the available levers for tactical control, give our solutions a wide applicability across many different domains of retail and e-commerce.

**Acknowledgements** We'd like to thank Tom Hall, Alison Oliver, and Priscilla Fearn for their support with this project, and Jason Zhang, Angelo Cardoso, and Bryan Liu for their helpful discussion.

# 8 APPENDIX

## 8.1 Inputs to Ithax algorithm

Table 6 shows example data used to construct for a hypothetical markdown event in time period $t + 1$. $P_{t+1}$ = {Product A, Product B, Product C}, whereas Product D $\notin P_{t+1}$ because $d_{productD}$=0. While cover and depth are defined at the product level, stock value and stock depth are defined over all products in the markdown. In this example, stock value=£2.7K, and stock depth=33%). We further note that the proposed solution for $t + 1$ allocates higher discount depths to worse-performing products (i.e. with higher cover).

## 8.2 Ithax helper sub-routine (is_adjustable)

The following algorithm is referred to in Algorithms 3-4 as "is_adjustable", and is used to identify the cover band $b_x$ that should be adjusted within the boundary adjustment step (Algorithm 3/4).

---
**Algorithm 5** Identify if a given band is adjustable

**Input:** $w$ is the minimum width for a cover band
**Input:** $b_x$ as in Algorithm 3
**Output:** Boolean for whether that band is adjustable or not

Initialise x = ($\frac{c_{b_x}^{max} - c_{b_x}^{min}}{2}$)
**return** ($depth_{b_x} > 0$) AND ($x \geq w$)

---

## 8.3 Algorithm 3: Example

In order to aid intuition, we show an example of the cover-depth mapping $B_i$ before and after a single operation of Algorithm 3, Table 7. In this example, target band $b_x$ = band (iv), and its upper bound $c_{iv}^{max}$ (Table 7, bold) is decremented by x=10 in order to halve $width_{iv}$. All higher bands with non-zero depth (bands (v)) are then adjusted to maintain their original widths (note: no gaps in cover between bands are permissible at any time). All affected boundaries that have been altered are italicised in the after adjustment column.

It is relatively easy to see how this set of adjustments might reduce the overall $M(\hat{P}_i)$: comparing before and after, the number of products at a relatively high depth (i.e. in band $b_x$) have been reduced as a proportion of the overall set of products. This may be counteracted by spikes in the density of products at certain levels of cover; while we do not observe a uniform level of products (and stock value) across cover, we generally did not need to adapt the algorithm to account for distributional concerns in order to achieve successful performance, because Algorithms 3 and 4 effectively work in tandem to reduce over- and under-shoots of $M^*$, to approach convergence (see Appendix 8.6 for more information).

*Note*: In Algorithms 3-4, band adjustments only proceed to $b_i^{highest}$. In Tables 7-9, we show adjustments as propagating through to the lower bound of the top band where depth=0, for clarity of intuition. We do not describe the adjustment of this top band in Algorithms 3-4, because it is of no functional importance.

## 8.4 Algorithm 4: Example

To aid intuition, we show two examples of the cover-depth mapping $B_i$ before and after band adjustment of different branches of Algorithm 4.

Option A In the first example (Table 8), we show how a single adjustment of the outer band adds more products to the highest depth banding, thus having the maximal impact on the stock depth with the change. This is congruent to the adjustments in lines 1-3 and line 9 in Algorithm 4. In this example, target band $b_x = b_{iii}$, and its upper bound $c_{iii}^{max}$ (Table 8, bold) is incremented by x=10 in order to increase the width of $b_x$. To ensure that there are no gaps in the bands, this change is then propagated up to $b_{iv}$ (i.e. by incrementing $c_{iv}^{min}$ by x=10). In doing so, we add more products to $\hat{P}$, and at a high depth, effectively increasing $M(\hat{P})$.

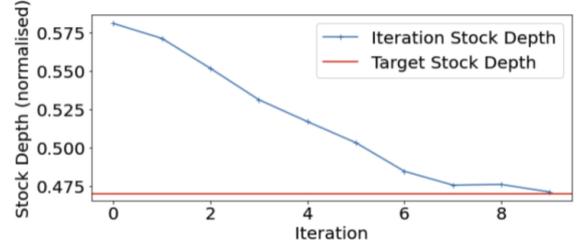

Figure 5: $M(\hat{P})$ by iterations of Ithax for a markdown event

Option B In the second example, Table 9 we show how adjusting one of the inner bands (i.e. where $b_x$ is not $b_i^{highest}$ or $b_i^{lowest}$) will lead to propagation of changes such that the change in magnitude to target band $b_x$ (denoted $x$ in Algorithm 4) leads to the highest possible impact on $M(\hat{P})$. In this example, target band $b_x = b_{ii}$, and its upper bound $c_{ii}^{max}$ (Table 9, bold) is decremented by x=5 in order to halve the width of $b_{ii}$. To ensure that there are no gaps in the banding the change is propagated up, to $b_{iii}$, but there does not need to be any changes in $b_i^{highest} = b_{iv}$ as the upper bound $c_{iii}^{max}$ doesn't decrease (as per line 24). This corresponds to the lines from line 16-24 of Algorithm 4.

It is relatively easy to see how both types of adjustments would increase the overall $M(\hat{P}_i)$. Comparing bands before and after adjustment in Tables 8 and 9, $b_x$ now contains the same or fewer products, but the highest bands in each table have increased in width. Any removal of products from lower target bands ($b_x$) is compensated for by the inclusion of more products at higher depths.

## 8.5 Convergence by Ithax algorithm

Figure 5 shows the stock depth over multiple iterations of boundary adjustment within Ithax (Algorithm 1). Stock value across all iterations remained within 0.1% of target. The convergence pattern shown is representative of Ithax's path, when in regular usage.

## 8.6 Assumptions and relevant considerations for Ithax's convergence

**Reasonable scaling of $V^*$, $M^*$** In developing Ithax, we assume that the target stock depth and stock values are scaled to a reasonable range. Stock depth targets in particular were expected to cover a 7-9%-point range, which had suited the needs of the business when markdown events were manually constructed by operations teams. Stock value targets were also assumed to be reasonably scaled, relative to the available stock value in the catalogue.

**Distributional assumptions regarding cover and stock value** One might expect based on Ithax's logic that solution quality might be modulated by the relationship between cover and stock value across the catalogue, or by the shape and density of the catalogue's cover distributions. We did not find it necessary to make any particular assumptions regarding the distribution of cover, or its relationship with stock value. However, it is certainly possible that these distributions may impact solution quality, and, for completeness, we report that our datasets did not generally see strong correlations between (a) cover and (product-level) stock value (generally, Pearson's r<0.25), or (b) cover and (product-level) full price (generally, Pearson's r<0.05). While we cannot report the cover distribution of our catalogue for commercial sensitivity reasons, we do report that it was generally bell shaped, with a long positive tail.

| Product | Sold Units at $t$ | Number of stock units at $t$ | Cover at $t$ | Proposed discount depth at $t+1$ | Full Price (£) | Discounted Price at $t+1$ (£) |
|---|---|---|---|---|---|---|
| A | 10 | 100 | 10 | 30% | 7 | 4.9 |
| B | 5 | 100 | 20 | 50% | 12 | 6 |
| C | 20 | 100 | 5 | 10% | 8 | 7.2 |
| D | 50 | 100 | 2 | 0% | 10 | - |

Table 6: Data with operational metrics defined for constructing a markdown event.

| Cover Band | Before adjustment $(c_{band}^{min}, c_{band}^{max}]$ | After adjustment $(c_{band}^{min}, c_{band}^{max}]$ | Discount Depth |
|---|---|---|---|
| i | (0, 20] | (0, 20] | 0% |
| ii | (20, 40] | (20, 40] | 10% |
| iii | (40, 60] | (40, 60] | 30% |
| [$b_x$] iv | (60, **80**] | (60, **70**] | 50% |
| v | (80, 90] | (70, 80] | 70% |
| vi | (90, ∞] | (80, ∞] | 0% |

Table 7: Single pass of Algorithm 3, adjusting $b_x$=band (iv)

| Cover Band | Before adjustment $(c_{band}^{min}, c_{band}^{max}]$ | After adjustment $(c_{band}^{min}, c_{band}^{max}]$ | Discount Depth |
|---|---|---|---|
| i | (0, 20] | (0, 20] | 0% |
| ii | (20, 40] | (20, 40] | 30% |
| [$b_x$] iii | (40, **60**] | (40, **70**] | 50% |
| iv | (60, ∞] | (70, ∞] | 0% |

Table 8: Single pass of Algorithm 4, expanding $b_x$=band (iii)

| Cover Band | Before adjustment $(c_{band}^{min}, c_{band}^{max}]$ | After adjustment $(c_{band}^{min}, c_{band}^{max}]$ | Discount Depth |
|---|---|---|---|
| i | (0, 20] | (0, 20] | 0% |
| [$b_x$] ii | (20, **40**] | (20, **30**] | 30% |
| iii | (40, 60] | (30, 60] | 50% |
| iv | (60, ∞] | (60, ∞] | 0% |

Table 9: Single pass of Algorithm 4, adjusting $b_x$=band (ii)

**Guidance regarding design of $B_0$** Ithax requires the user to specify an initial cover-depth mapping, $B_0$, which serves as the starting point for Ithax. If $B_0$ is chosen favourably (i.e. is in line with targets), we have often found that only a single pass through the algorithm results in successful convergence (i.e. all bands are only adjusted once). Below, we list some assumptions and guidance regarding the design of $B_0$ that encourage faster convergence:

- Maximum stock depths: We assume that the highest depth in $B_0$ ($d_b^{highest}$) has to be greater than the stock depth target $M^*$. This is a straightforward assumption as if this didn't hold, there would be no way to hit the target with the depths present in $B_0$.
- Minimum band widths: In order to allow Ithax to hit a wide variety of targets, we generally found it preferable to design $B_0$ where each band had $width_b > 4w$, where $w$ refers to the minimum width threshold used ($w$=3, in our case).
- Overshoot first: We generally found that it was preferable to design $B_0$ such that the first iteration of Ithax would find $M(\hat{P}_i) > M^*$ (i.e. triggering Algorithm 3). This is usually reasonable to achieve, as the target $M^*$ was generally within the lower range of included depths, and much smaller than $d_b^{highest}$. We expected this to be the case because in general because working to increase $M(\hat{P}_i)$ in the first iteration involved more edge cases with regard to increasing $c_{b^{highest}}^{max}$ and potentially reaching a range of cover in which there were not many products in $\mathfrak{P}$.
- Reaching the lowest band without convergence: One of the other implementation concerns to note is the algorithm "bottoming out" in Algorithm 3 and 4: reaching the lowest band without convergence. In practice we found that this was easily solved by adjusting $B_0$ such that the intervals remain wide enough for each individual band that the algorithm doesn't encounter this situation. In practice, adjusting $B_0$ allowed us to avoid non-convergence every time we encountered this situation in regular usage.